\documentclass[12pt]{article}

\usepackage{scicite,soul,xcolor}
\usepackage{amsmath}
\usepackage{dcolumn}
\usepackage{bm}
\usepackage{times}
\usepackage{graphicx} 

\usepackage{hyperref}

\newcommand{\dw}{\boldsymbol{w}}
\newcommand{\bF}{\boldsymbol{F}}
\newcommand{\bxi}{\boldsymbol{\xi}}
\newcommand{\bA}{\boldsymbol{A}}

\newenvironment{sciabstract}{%
\begin{quote} \bf}
{\end{quote}}

\title{Robust control for multi-legged elongate robots in noisy environments}

\author
{Baxi Chong$^{1,\dag}$, Juntao He$^{2,\dag}$, Daniel Irvine$^{3}$, Tianyu Wang$^{2}$, Esteban Flores$^{4}$ \\ Daniel Soto$^{2}$, Jianfeng Lin$^{1}$, Zhaochen Xu$^{1}$, Vincent R Nienhusser$^{4}$\\ 
Grigoriy Blekherman$^{3}$, Daniel I. Goldman$^{1,2,\ast}$\\
\\
\normalsize{$^{\dag}$ These authors contributed equally to this work.}\\
\normalsize{$^{1}$School of Physics, Georgia Institute of Technology,}\\
\normalsize{837 State St NW, Atlanta, GA 30332, USA}\\
\normalsize{$^{2}$Institute for Robotics and Intelligent Machines, Georgia Institute of Technology,}\\
\normalsize{801 Atlantic Dr NW, Atlanta, GA 30332, USA}\\
\normalsize{$^{3}$School of Mathematics, Georgia Institute of Technology,}\\
\normalsize{686 Cherry St NW, Atlanta, GA 30332, USA}\\
\normalsize{$^{4}$Department of Mechanical Engineering, Georgia Institute of Technology,}\\
\normalsize{801 Ferst Dr NW, Atlanta, GA 30318, USA}\\
\\
\normalsize{$^\ast$To whom correspondence should be addressed; E-mail: daniel.goldman@physics.gatech.edu}
}

\date{}

\begin{document} 


\baselineskip24pt


\maketitle


\begin{sciabstract}







Modern two and four legged robots exhibit impressive mobility on complex terrain, largely attributed to advancement in learning algorithms. However, these systems often rely on high-bandwidth sensing and onboard computation to perceive/respond to terrain uncertainties. Further, current locomotion strategies typically require extensive robot-specific training, limiting their generalizability across platforms.  Building on our prior research connecting robot-environment interaction and communication theory, we develop a new paradigm to construct robust and simply controlled multi-legged elongate robots (MERs) capable of operating effectively in cluttered, unstructured environments. In this framework, each leg-ground contact is thought of as a basic active contact (bac), akin to bits in signal transmission. Reliable locomotion can be achieved in open-loop on ``noisy” landscapes via sufficient redundancy in bacs. In such situations, robustness is achieved through passive mechanical responses. We term such processes as those displaying mechanical intelligence (MI) and analogize these processes to forward error correction (FEC) in signal transmission. To augment MI, we develop feedback control schemes, which we refer to as computational intelligence (CI) and such processes analogize automatic repeat request (ARQ) in signal transmission. Integration of these analogies between locomotion and communication theory allow analysis, design, and prediction of embodied intelligence control schemes (integrating MI and CI) in MERs, showing effective and reliable performance (approximately half body lengths per cycle) on complex landscapes with terrain ``noise” over twice the robot’s height. Our work provides a foundation for systematic development of MER control, paving the way for terrain-agnostic, agile, and resilient robotic systems capable of operating in extreme environments.

\end{sciabstract}



\section*{Introduction}


Locomotion is essential to not only living systems~\cite{alexander2003principles,childress2012natural} but also engineered devices such as robots~\cite{gravish2018robotics,kim2013soft,Aguilar:2016bq}. Specifically, legged robotic systems are an important paradigm for locomotion on rough/cluttered terrains, partially because of their capability to schedule contact points with the environment and overcome obstacles comparable to their leg sizes~\cite{weber1836mechanik,lee2020learning}. For bipedal and quadrupedal robots, designing agile locomotion has been a longstanding research problem~\cite{raibert1986legged}. The difficulties lie in developing appropriate robot intelligence to detect and respond to terrain uncertainties (e.g., maintaining stability while performing self-propulsion on complex terrains)~\cite{tan2018sim}. Computational intelligence (CI) - sensing, processing, and responding via active feedback controls - is widely recognized as a crucial component of robot intelligence~\cite{raibert2008bigdog,hutter2016anymal}.	Recent advances in machine learning, sensing, and control algorithms have significantly enhanced computational intelligence, enabling agile and adaptive behaviors in bipedal and quadrupedal robots~\cite{rudin2022learning,huang2023efficient}.


In addition to CI, researchers have considered other forms of robot intelligence, such as mechanical intelligence (MI): the passive response to environmental perturbations governed by physical laws or mechanical constraints~\cite{sitti2021physical,wang2023mechanical}. Of our particular interest, researchers considered increasing the number of legs and developed multi-legged robots for which additional legs could improve the static stability and therefore help avoid catastrophic failures (e.g., loss of stability). Early examples include RHex, with six legs attached to a rigid body~\cite{saranli2001rhex}, designed for stable locomotion over uneven terrains. More recently, we developed multi-legged elongate robots (MERs) using serially connected bipedal units, enabling self-reconfiguration to maintain stability in extreme environments~\cite{ozkan2020systematic,chong2022general,chong2023self}. 

Beyond stability, our prior work~\cite{chong2023multilegged} further quantified the benefits of locomotion reliability emerged from leg redundancy in MERs. Specifically, we developed an analogy between multi-legged  locomotion on rugose terrain and signal transmission over noisy channel in communication theory~\cite{chong2023multilegged}, where ``message” being transmitted is the body of the robot\footnote{The bit in the message is similar to legs of a robot generating thrust from interaction with terrains.} and ``coding" is the internal coordination/synchronization among the many legs (Table 1). In this analogy, CI and MI in multi-legged locomotion correspond to automatic-repeat request (ARQ, where errors are detected and a re-transmission is requested) and forward error correction (FEC, where errors are corrected via redundancy and coding algorithms without re-transmission), respectively, in signal~transmission~\cite{pierce2012introduction}. This analogy allowed us to demonstrate that with sufficiently many legs, it is possible to generate effective open-loop locomotion on rugose terrains with guaranteed reliability. 



Despite the robustness, the mechanically intelligent MERs from our prior work were still far from practical applications for two key reasons. First, those MERs are substantially slower compared to other legged robots. Second, achieving the desired level of robustness often requires excessively many legs (e.g., 16 legs in~\cite{chong2023shannon}), which complicates the design and limits practicality. We hypothesize that these challenges can be addressed by either (1) tailoring MI to specific locomotion environments or (2) incorporating CI into those already mechanically intelligence robots. 

For (1), robot error-correction through MI can emerge from distinct mechanisms. For example, robots with redundant legs can benefit from global passive responses acting on the body’s center of gravity (e.g., rigid-body pitch, yaw, and roll), analogous to ``message"-level error correction in signal transmission. Additionally, legs with specific geometry and compliance, such as the C-shaped legs used in RHex designs~\cite{moore2002leg,jun2014compliant}, can buffer local interactions with rugose terrain, analogous to error correction at the ``bit"-level. These observations raise fundamental questions: (i) How can MI, such as leg number and geometry, be adapted to different environments? (ii) How can we control robots with varying forms of MI? (iii) Is there a fundamental upper bound on speed for a given MI design? For (2), similar to hybrid ARQ algorithms in communication theory, which integrate FEC and ARQ towards more efficient signal transmission, it is intriguing to explore how the integration of MI and CI can enhance MER locomotion performance, advancing toward a more unified form of embodied intelligence.  Considering the CI, there is also a lack of intuition behind (iv) what to sense and (v) how to respond for multi-legged locomotion.

\begin{table} 
	\centering
	\caption{\textbf{Analogy between communication theory and robot locomotion} }
	\label{tab:example} 
	\vspace{1em}
	\begin{tabular}{| p{7cm} | p{7cm} |} 
		\hline
		\textbf{Communication theory concepts} & \textbf{Analogy in Locomotion} \\
		\hline
        \hline
		\textit{Signal} & The robot self-transport towards destination \\
        \hline
		\textit{Bit} & Bac (basic active contact), such as a step in legged locomotion \\
        \hline
        \textit{Coding} & The algorithm to control the robot locomotion \\
        \hline
		\textit{Channel noise} & Terrain complexity which introduce noises in the  bac pattern \\
        \hline
		\textit{Encoding} & (Spatial and temporal) coordination among the many bacs\\
        \hline
		\textit{Decoding} & Response to channel noise and its effect on performance \\
		\hline
        \textit{Forward Error Correction (FEC)} & Mechanical intelligence: passive (mechanical) response to noise\\
        \hline
        \textit{Automatic-Repeat Qequest (ARQ)}  & Computational intelligence: active response (feedback control) to noise\\
        \hline
        \textit{Hybrid ARQ} & Embodied intelligence: integrated mech. and comp. intelligence\\
        \hline
        \textit{Source entropy}  & The maximum possible transportation rate in noise-free conditions\\
        \hline
        \end{tabular}
        \newline
        \vspace{1em}
        \newline
        
        \begin{tabular}{| p{7cm} | p{7cm} |} 
        \hline
        \textbf{Example algorithm} & \textbf{Analogy in Locomotion} \\
        \hline
        \hline
        \textit{Majority vote} & Passive response by gravity to redistribute bacs favorably\\ 
        \hline
        \textit{Simple repetition} & Synchronizing multiple legs to produce identical thrust \\   
        \hline
        \textit{Majority vote window size} & Spatial period in spatial synchronization; or uniformity of thrust generation function in temporal synchronization \\ 
        \hline
        \textit{Adaptive majority vote} & Adaptive vertical wave with linear feedback control\\ 
        \hline
	\end{tabular}
\end{table}



To address these questions, we use geometric mechanics as a tool to analyze and design MI/CI in MERs. Geometric mechanics (GM), or gauge kinematics~\cite{wilczek1989geometric,shapere1989gauge}, has been developed~\cite{shapere1987self,berry1990anticipations,montgomery1990,marsden1990reduction,krishnaprasad1994,kelly1995geometric,ostrowski1998geometric,lewis2000,Bloch2003,melli2006} since 1980s as a general framework to connect locomotor performance to arbitrary patterns of self-propulsion. With diagrammatic plots, GM offers quantitative insight into how unconventionally-shaped robots can generate self-propulsion patterns to obtain desired behaviors. Here, we use GM to explore how adding more legs can enhance speed by optimizing the encoding of MI (Table 1). Specifically, we show that additional legs can contribute to locomotion speed only when coupled with active body undulation. The undulation generates a propagating wave that travels from head to tail. The wave propagation rate and amplitude regulate the trade-off between locomotion speed and robustness. Finally, using a comparative theoretical, numerical, and experimental approach, we illustrate that there exist an upper bound of MER locomotion on flat ground. 

To approach this theoretical speed limit on complex terrains, we design CI control schemes that enhance speed while preserving robustness in locomotion. Using simple contact sensors to estimate terrain rugosity, we develop two CI frameworks that enable robots to adaptively shift along the robustness-speed spectrum. These CI-based strategies yield effective locomotion across noisy landscapes, approaching the predicted upper speed bound. Alternatively, we show that modifying leg morphology, such as equipping MERs with C-shaped legs inspired by RHex~\cite{moore2002leg,jun2014compliant}, offers a complementary MI-based approach tailored to specific terrain, also achieving performance close to speed upper bound. Collectively, these results show that abstract principles derived from Shannon’s communication theory can be concretely applied to enhance the locomotion capabilities of multilegged robots.

\section*{Results}

\subsection*{Gravity as an MI control scheme}

We first investigate how motors/sensors should be arranged such that one can guarantee that a robot can go from point A to point B in a specified time across a noisy landscape. This question is similar to signal transmission over noisy channels as analyzed by Shannon nearly a century ago~\cite{shannon1948mathematical,moore1956reliable}. To counter channel noise, Shannon~\cite{shannon1948mathematical} constructed a scheme, in which the central idea was to digitize (encode) information into bit sequences and ``approximate" (decode) the channel-contaminated bit sequences via redundancy and appropriate algorithms. ``Simple repetition” is one of the oldest and simplest encoding schemes for signal transmission, where the same bit sequence is repeated multiple times before being sent over a noisy channel~\cite{pierce2012introduction}. The corresponding decoding scheme is ``majority vote,” where the receiver estimates the original bit based on the most frequently occurring value in the received repeated sequence. Previous research has demonstrated that a simple repetition encoder together with majority vote decoder can be sufficient to ensure reliable signal transmission over a noisy channel~\cite{shannon1948mathematical}. 

Similar to the bit-based digital signal transmission, we consider a dissipation-dominated system in which locomotion is driven by thrust generation from basic active contacts (bacs, our analogy to bits). We quantify the relative importance of inertia using the coasting number ($\mathcal{C}$), a dimensionless parameter analogous to the Reynolds number in fluid mechanics~\cite{rieser2024geometric}. In our system, we find that $\mathcal{C} \approx 0.01 \ll 1$, indicating that locomotion is thrust-dominated, with inertia playing a negligible role in movement dynamics.  In multi-legged locomotion, a bac typically corresponds to a single step. If a bac is perturbed by the environment (e.g., a missing step), gravity will rapidly redistribute the supporting forces to balance the system. With proper coordination, the effect of bac noise will be shared among all bacs instead of acting on an individual bac. This effectively leads to a moving average filter over the perturbation (Fig.~1C.1). Such a mechanical moving average filter shares a analogous action as the majority vote decoder in signal transmission. Moreover, the response to gravity is a passive process without additional computational power (e.g., on sensors and processors). Thus, such gravity-induced passive processes can be categorized as a MI control scheme.

To coordinate with the majority vote passive decoder, it is natural to control  (``encode") the MER locomotion with ``simple repetition". That is, the bacs are spatially repeated. Specifically, each module, where a module is chosen to be a bipedal robot, performs a synchronized thrust generation. That is, all legs simultaneously make contact with the ground, generating identical instantaneous thrust during the bac, and simultaneously disengage from the ground at the end of the bac. The additional legs (bacs) in this configuration primarily contribute to providing redundancy in thrust generation rather than generating additional speed~\cite{chong2022general}. With spatially repeated bac and gravity-induced MI control (``majority vote" decoder), prior work illustrated the effectiveness of using redundant legs to counter the noises from environmental perturbation (Fig.~2.B.1). To test the prediction, the time required to travel 60 $cm$, $T_{[D=60]}$, was recorded for MERs with different number of legs on flat (Fig.~2.B.1, blue curve) and rugose terrains (green curve).  Experiments suggested that adding more legs not only improves the average terrain-disturbed average speed, but also reduces the variance.

\subsection*{Spectrum between speed and robustness}

Despite the intrinsic robustness of MERs in complex terrain, prior work also suggested that the speed in noise-free environments does not scale with the increase in leg number, leading to reduced relative speed (normalized by body length). We posit that the limited absolute speed can be attributed to the simple repetition encoding (additional legs performing repetitive thrust generation tasks). In analogy with communication theory (appropriate algorithms to substantially improve signal transmission rate compared with the simple repetition), we expect that appropriate encoding schemes can substantially improve locomotion performance in noise-free environments. 

We first test a series of \textit{ad-hoc} control schemes. Specifically, we quantify control redundancy as $S_n$, prescribing how many legs are instantaneously performing the same thrust generation task. Following this definition, $S_n$ is similar to the window size in the ``majority vote" decoder. The detailed self-deformation pattern prescription can be found in SI (Sec. 2). Here, we illustrate snapshots of a 12-legged MER implementing $S_n=1$ and $S_n=3$ in Fig.~2.A.2. Similarly, we recorded $T_{[D=60]}$ for a series of $S_n$ on flat (blue curve) and rugose (green curve) terrains in Fig.~2B.2. We notice that decreasing $S_n$ will improve the speed in noise-free environment but with a cost in robustness to complex terrains, indicating that \textit{there exists a spectrum between speed and robustness, modulated by the encoding schemes}. 

\subsection*{Encoding with geometric mechanics}

To achieve longer stride lengths in robots, engineers typically use longer legs instead of more legs~\cite{full1999templates}. In other words, it is intuitive that the stride length scales linearly with the leg length. But it remains unclear how the stride length should scale with the number of legs. In this section, we will use geometric mechanics to ``encode" multi-legged locomotion and explore mechanisms to increase stride length in a systematic way.

While the scaling problem is an open question for legged locomotion, we can take valuable insights from the limbless locomotion literature in biology~\cite{rieser2019geometric,chong2022coordinating,wang2023mechanical}. Studies have shown that elongate limbless organisms across different scales, from microscopic nematode worms to macroscopic snakes/lizards, achieve a similar relative speed of approximately 0.4 body lengths per cycle through similar patterns of body undulation. Inspired by this, we posit that addition of legs can contribute to speed with active body undulation. 

To establish a model of body undulation, we first consider an abstract characterization of locomotion of a unit bipedal robot (with one pair of legs). The periodic lifting and landing of contralateral\footnote{The pair of legs connected to the same body segment.} legs together with the leg shoulder retraction/protraction offer the capability to generate thrust (bac) and therefore locomote. Specifically, during locomotion, each leg unit periodically undergoes a retraction phase while in contact with the ground (backward stroke) and a protraction phase when it disengages from the ground. The detailed bac prescription can be found in SI. Sec.2. In this way. the capability of a unit bipedal robot can then be characterized by the leg length ($l$) and the shoulder angle amplitude ($A_{leg}$). 

Next, we use actuated yaw joints to connect the consecutive pairs of legs and enable a wave of lateral body undulation. As documented in prior work~\cite{chong2022general,chong2023self}, we assume that the lateral body undulation wave can be prescribed by:

\begin{align}
    \alpha(i,t) &= w_1(t)\beta_1(i) + w_2(t)\beta_2(i)  \nonumber \\ 
    \beta_1 (i) &= \sin{(2\pi S_n \frac{i}{N})}  \nonumber \\ 
    \beta_2 (i) &= \cos{(2\pi S_n \frac{i}{N})} 
\end{align}\label{eq:shapebasisfunc}

\noindent where $\alpha(i,t)$ denotes the yaw angle of $i$-th joint at time $t$; $\beta_1(i)$ and $\beta_2(i)$ are shape basis functions (SI. Sec. 2.1); $N$ is the number of leg pairs; $w_1(t)$ and $w_2(t)$ are the time series of weights for the corresponding shape basis functions. Further from our results in Fig.2.B.2, we set $S_n=1$ unless otherwise noted to obtain longer stride. Here, we define the shape variable as $\dw(t) = [w_{1}(t), w_{2}(t)]$, which then uniquely characterizes the posture of body movement. As discussed in prior work ~\cite{chong2023self,chong2022general}, the contact pattern of legs and the leg shoulder angles can also be prescribed by the shape variable $\dw$ (detailed prescription equations can be found in SI Sec. 2). In Fig.~3.A, we illustrated the shape space for a 12-legged MER with $l=8.4\ cm$  and $A_{leg}=\pi/12$. Notably, some regions in the shape space is inaccessible in our current robot implementation because of self-collision among adjacent legs (e.g., Fig.~3.A.1).

Perturbation in the shape variables (self-deformation) can result in net displacement (self-propulsion). For example, in Fig.~3.A, the self-deformation (form the posture circled in red to the posture circled in black) can cause net translation and rotation in the world reference frame. We quantify net translation and rotation with three variables: $\Delta x,\ \Delta y,\ \Delta \theta$, in forward, lateral, and rotational directions respectively (Fig.~3.A.3).

We define the body velocity, $\bxi=[\xi_x, \xi_y, \xi_\theta]$, as the instantaneous  locomotor velocity in the forward, lateral, and rotational directions~\cite{murray1994mathematical}.  Reference to Fig.~3.A.3, we have: $\bxi=\lim_{t\rightarrow0} \frac{[\Delta x,\ \Delta y,\ \Delta \theta]}{dt}$. We can then numerically calculate the body velocity from shape variables ($\dw$) and the shape velocity ($\dot{\dw}$) via net ground reaction forces (GRF) analysis. Specifically, the GRF experienced by the robot is the sum of the GRF experienced by all legs in stance phase\footnote{We consider a leg to be in stance phase if it is in contact with the ground.}. On flat hard ground with dry isotropic friction, the GRF ($F$ in Fig.~3.B) should have a fixed magnitude in the direction opposite to the direction of slipping ($v$ in Fig.~3.B). From geometry, the direction of slipping at each foot can be uniquely expressed as a function of the shape variable ($\dw$), shape velocity ($\dot{\dw}$), and body velocity ($\bxi$). Assuming quasi-static motion in the low coasting system~\cite{rieser2024geometric}, we consider the total net force applied to the system is zero at any instant in time:

\begin{equation}\label{eq:forceIntegral}
    \bF=\sum_{i\in I_{\dw}} {\left[\bF^{i}_{\parallel}\left(\bxi,\dw,\dot{\dw}\right)+\bF^{i}_{\perp}\left(\bxi,\dw,\dot{\dw}\right)\right]}=0,
\end{equation}

\noindent where $I_{\dw}$ is the collection of all stance-phase legs, determined by the shape variable $\dw$. At a given body shape $\dw$, Eq.(\ref{eq:forceIntegral}) connects the shape velocity $\dot{\dw}$ (e.g., Fig.~3.A.2) to the body velocity $\bxi$. Therefore, by the implicit function theorem and the linearization process, we can numerically derive the fundamental equation of motion:

\begin{equation}
    \bxi ~\approx \bA(\dw)\dot{\dw} = \begin{bmatrix}
        \bA^x(\dw) \\ \bA^y(\dw) \\ \bA^\theta (\dw)
    \end{bmatrix}\dot{\dw} ,
\end{equation}

\noindent where $\bA$ is the local connection matrix, $\bA^x, \bA^y, \bA^\theta$ are the three rows of the local connections respectively. Each row of the local connection matrix can be regarded as a vector field over the shape space, called the connection vector field (Fig.~3.C.3). In this way, the body velocities in the forward, lateral, and rotational directions are computed as the dot product of connection vector fields and the shape velocity $\dot{\dw}$. 

Consider a gait as a periodic pattern of self-deformation: $\{\partial \phi\ : [w_1(t), w_2(t)], t\in(0,T]\}$, where $T$ is the temporal period. The displacement along the gait path $\partial \phi$ over a cycle can be approximated to the first order by:

\begin{equation}\label{eq:lineintegral}
    \begin{pmatrix} 
        \Delta x \\
        \Delta y \\
        \Delta \theta 
    \end{pmatrix}
    \approx  \int_{\partial \phi} {\begin{bmatrix}
        \bA^x(\dw) \\ \bA^y(\dw) \\ \bA^\theta (\dw)
    \end{bmatrix}\mathrm{d}\dw}.
\end{equation}

Of our particular interest to improve the stride length ($\Delta x$), we have $\Delta x \approx \int_{\partial \phi} \bA^x(\dw) \mathrm{d}\dw$. According to Stokes' Theorem, the line integral along a closed curve $\partial \phi$ is equal to the surface integral of the curl of $\bA^x(\dw)$ over the surface enclosed by $\partial \phi$:

\begin{equation}\label{eq:stokes}
    \Delta x \approx \int_{\partial \phi} {\bA^x(\dw)\mathrm{d}\dw}=\iint_{\phi} {\nabla\times \bA^x(\dw)\mathrm{d}w_1\mathrm{d}w_2},
\end{equation}

\noindent where $\phi$ denotes the surface enclosed by $\partial \phi$. 
The curl of a connection vector field, e.g., $\nabla\times \bA^x(\dw)$, is referred to as a height function~\cite{hatton2015nonconservativity}. Specifically, $\nabla \times \bA^x(\dw)$ represents the forward height function, which quantifies how self-deformation generates a net translation in the forward direction. With the above derivation, the gait design problem is simplified to drawing a closed path in the shape space. Net displacement over a period can be approximated by the integral of the surface enclosed by the gait path. Hence, we are able to visually identify the optimal gait leading to the longest stride by finding the path with the maximization surface integral. 

We illustrate the forward height function and an example gait path in Fig.~3.C.4. For simplicity, we only consider the circular gait path in the shape space: $w_1(t)=A_b\sin(t),\ w_2(t) =A_b\cos(t)$, where $A_{b}$ sets the magnitude of body undulation. From the structure of the forward height functions, we notice that (i) 
at low amplitudes, increasing $A_{b}$ can almost linearly increase the stride, (ii) there exists a threshold $A_b^*$ beyond which further increasing $A_{b}$ can enclose volumes of opposite signs, leading to shorter stride, and (iii) optimal $A_{b}$ could be infeasible because it will lead to self-collision among different legs. We verify our predictions using robot experiments (Fig. 3.C.4). We illustrate snapshots of a MER implementing $A_b = 0$ and $A_b  =  \pi/6$ in Fig.~3.C.2.

\subsection*{Emergent upper bound of absolute speed with additional legs}

We next use geometric mechanics to develop an encoding scheme for MERs with different numbers of legs. Following similar processes, we numerically obtained the forward height functions and tested the modeling predictions on the robot experiments. We illustrated forward height functions and experimental results with $N=7$ and $N=3$ in Fig.~4. We also labeled the infeasible region in the shape space (due to self-collusion) in the left panels. Here, we define the self-collision as the body configurations in which the distance between any adjacent feet is less than a 1~$cm$ (approximately the size of the robot feet). We defined $A_{SC}$ as the critical body undulation amplitude above which the substantial self-collision will occur. Similarly, we define $A_b^*$ as the optimal body undulation amplitude leading the longest stride (ignoring the self-collision). 

For a given multi-legged system, we choose the appropriate $A_b$ with the longest stride without self-collision using the following equations:

\begin{equation}
        A_b(N)= 
\begin{cases}
    A_{SC},& \text{if}\ \  A_{SC}<A_b^*\\
    A_b^*,   & \text{if}\ \ A_{SC}>A_b^*\\
\end{cases}
\end{equation}

We also notice that on the one hand, $A_b^*$ will decrease as we increase $N$. In other words, we need a lower amplitude body undulation to achieve optimal stride as we increase the number of legs. On the other hand, $A_{SC}$ will increase as we increase $N$. That is, we can afford greater body undulation amplitude without causing self-collision as we increase the number of legs. In our robot experiments, we notice that $A_{SC}<A_b^*$ when $N<7$; and $A_{SC}>A_b^*$ when $N>7$. Based on the observation, we hypothesized that increasing the number of legs can lead to substantially longer stride when $N<7$; however, above $N=7$, further increasing the number of legs will not lead to longer stride. 

We verify our hypothesis in numerical simulation and robot experiments.  Both analysis suggest that there exists a $N$ above which adding more legs will not contribute to absolute speed (Fig.~4.B.1). Further, the saturation $N$ in both numerical and experimental analysis is reasonably close to 7, as predicted in theory. Finally. there is an emergent upper bound of absolute speed (approximately 33 $cm$/cycle) no matter how many legs we could add to the MER. Notably, the number 7 is subject to our choice of leg length, shoulder angle amplitude ($A_{leg}$), and spatial period ($S_n$). We predict that we can achieve saturated speed with fewer number of legs if we can have  longer legs, larger shoulder angle amplitude, or lower spatial period.

\subsection*{Source entropy of multi-legged locomotion}

Notably, the emergent upper bound of the absolute speed in noise-free environments shares similar concepts to the source entropy in communication theory (Table 1). In this section, we will justify the existence of such an upper bound using two simple kinematic models. 

First, we use a non-slip model, where we assume no slipping can occur between each foot and the substrate. We also assume that one of the contralateral legs must always be in contact with the substrate (i.e., no locomotor aerial phases). In this case, from geometry, we can estimate the upper bound of the stride length to be $4l$, where $l$ is the leg length (Fig.~4.C.1).

We then consider the slipping between the feet and the substrate. For each foot, we define $d_1$ as the total amount of slipping in forward direction, and $d_2$ as the total amount of slipping in backward direction. Assuming the isotropic Coulomb dry friction, the net kinetic energy that the robot obtains per cycle can be calculated as $(d_1-d_2)F$, where $F$ is the frictional force~\cite{chong2023self}. In steady states, where we do not expect changes in kinetic energy over cycles, thus, we have $d_1=d_2$. In this case, the upper bound of the stride remains to be $4l$ (Fig.~4.C.2).

With our body undulation encoding scheme, the saturated stride ($33\ cm/cycle$) is reasonably close to the predicted upper bound ($4l=33.6\ cm/cyc$) at $N=7$ (black dashed line in Fig.~4.B.1). That being said, with better encoding scheme, we speculate that it is possible to reach the predicted upper bound with fewer legs.

\subsection*{Temporal and spatial synchronization for simple repetition}

Despite the effectiveness in noise-free environments (Fig.~5.B. blue curve), the benefit of using lateral body undulation is substantially compromised when MERs move on rugose surfaces (Fig.~5.B. green curve). Inspired by communication theory, we further explore other encoding mechanisms to simultaneously improve the robustness and the speed over terrain complexity. 

In prior work~\cite{chong2023multilegged}, we demonstrated that the contact modulation by the addition of a vertical wave of body movement can offer robustness over contact errors from terrain rugosity. Yet, it remains unclear what mechanism drives such robustness and more importantly, how to adaptively modulate the vertical wave to terrains with different rugosity. To generate a well-defined vertical wave, we include a pitch joint in addition to the yaw joint to connect two bipedal modules. Specifically, the pitch joints are prescribed by:

\begin{equation}
    \alpha_p (i,t) = A_p \cos{\big(2\times (2\pi S_n\frac{i}{N}+t)\big)},
\end{equation}

\noindent where $\alpha_p(i,t)$ denotes the pitch angle of i-th joint at time $t$, $A_p$ is the magnitude of vertical wave. We illustrate an example of a vertical wave in Fig.~5.C. Notably, without the vertical wave, the contralateral legs have the opposite contact state, making the average bac duration to be $\tau=T/2$, where $T$ is the temporal period. The use of vertical wave systematically reduce the bac duration of each leg, where $\tau$ decreases as $A_p$ increases.

We investigate the mechanisms to improve robustness via vertical wave contact modulation.  As discussed before, the gravity-induced passive decoder shares a similar mechanism with the ``majority vote", which is the most effective when coupled with simple repetition encoding where all legs are doing exactly the same task. In other words, let $f(t,i)$ be the instantaneous thrust generated by $i$-th leg at time $t$. The thrust is defined as the projection of GRF, $F$ in Fig. 3, to the direction of motion. For simple repetition, we want to achieve $f(t,i) = f(t,j), \ \ i,j\in I_m$, where $I_m$ is defined in Eq.~2.

Spatial synchronization is a straight-forward way to implement ``simple repetition". That is, all the legs are synchronized at every instant such that $f(t,i) = f(t), i\in I_m$. Intuitively, the ``window size" of the ``majority vote" in spatial synchronization is characterized by the spatial redundancy (the number of legs)~\cite{chong2023shannon}. Here, we consider the temporal synchronization for ``simple repetition" by reducing the temporal variation in the thrust generation within a bac duration. That is, we aim to achieve $f(t,i)=f(i), t\in\tau$ where each bac will generate time-invariant thrust. 

Formally speaking, we define $\hat{v}$ as the stride length on rugose terrains over one cycle. We characterize the locomotion robustness by the coefficient of variation ($C_v$, ratio of the standard deviation to the mean) in $\hat{v}$. In principle, a constant thrust generation function $f(t,i)=f(i), t\in\tau$ is a minimizer of $C_v(\hat{v})$ (proof in SI). However, in practice, we cannot directly engineer $f(t,i)$ and reinforce the constant thrust generation function. Instead, we show that a more uniformly distributed thrust generation function (characterized by the variance of $f(t,i)$, defined in SI) can result in lower $C_V(\hat{v})$ (proof in SI).

As illustrated in Fig.~5.C, a vertical wave can serve to reduce the variance of thrust generation function, $f(t,i)$. In this way, the use of vertical wave will contribute to temporal synchronization. In this way, the ``window size" of the ``majority vote" in temporal synchronization is characterized by the uniformity of the thrust generation function. On the other hand, the vertical wave reduces the bac duration $\tau$, and effectively reduce the average number of legs in stance phase at each moment.  Because of the two competing factors on robustness with the vertical wave, we hypothesize that there exists an optimal vertical wave amplitude, $A_p^*$, that will lead to the most reliable locomotion (characterized by the lowest $C_V(\hat{v})$). Further, $A_p^*$ should be a function of terrain complexity. 

To test our predictions, we numerically computed the CDF of average terrain-disturbed average speed for $N=6$ serially connected modules over 10 periods subject to different levels of vertical wave modulation. We notice that appropriate vertical wave (systematically shortening $\tau$ to 0.4T) can indeed lower the variance on the average terrain-disturbed velocity. However, further increasing the vertical wave amplitude (systematically shortening $\tau$ to 0.3T) can result in higher variance on the average terrain-disturbed velocity because of the excessive total contact loss.

We then test our predictions on robot experiments. We constructed a model rugose terrain composed of ($10\times10$ cm$^2$) blocks with variation in heights. The block heights, $h(x,y)$, are randomly distributed (SI). We define the terrain rugosity, $R_g$, as the standard deviation of heights normalized by block side length. We tested the performance of a 12-legged MER on the flat ($R_g=0$) and the rugose ($R_g=0.17$) terrains subject to different magnitude of vertical wave ($A_{p}$). We measured the average velocity (over 10 trials, each trial contains 3 cycles) and plotted it against $A_{p}$. We notice that appropriate vertical wave amplitude can cause the most reliable and predictable (indicated by the variance) locomotion. 

To further illustrate the importance of choosing the appropriate $A_{p}$, we experimentally obtained the empirical CDF of cycle-average velocity of a 12-legged MER moving on the rugose terrains subject to different magnitude of vertical wave (Fig.~5.F). We illustrate that the optimal $A_{p}$ is approximately $\pi/9$: deviation from this optimal amplitude in either direction can result in higher variance in velocity. 

\subsection*{CI in MER locomotion: bac error detection and correction}

Thus far, we have demonstrated that appropriately designed MI can enable effective locomotion on noisy landscapes. In particular, by choosing appropriate coding schemes, it is possible to trade off between locomotion speed and robustness. However, to simultaneously enhance both speed and robustness, additional strategies may be required. According to the no-free-lunch theorems~\cite{wolpert1997no}, it may be overly ambitious to optimize both objectives using MI alone. We posit that incorporating computational intelligence (CI) can overcome this limitation. Nevertheless, for MERs with high-dimensional control parameter spaces, identifying which variables to sense and determining the appropriate system responses becomes increasingly challenging.

One of the most straightforward forms of CI is to sense every bac and immediately apply a correction. That is, when a bac is missed, the system immediately attempts to correct it, such as by firing an repeating bac. In this scheme, bac error detection and correction are applied at the level of individual bacs. This catch-and-correct mechanism is analogous to the Stop-and-Wait ARQ protocol~\cite{tanenbaum1981network}, in which the sender transmits one unit of data at a time, and the receiver checks for transmission errors; if an error is detected, a retransmission is requested via the feedback channel.

Despite its simplicity, such catch-and-correct mechanisms can be inefficient. Inspired by communication theory, where more advanced algorithms have been developed beyond the Stop-and-Wait ARQ, we explore more advanced CI in MER, in which bac detection and correction are applied at the level of the entire robot, rather than at individual bacs.

\subsection*{Single-input-single-output CI}

We note that average rugosity serves as a measure of the global complexity of the terrain. However, particularly for terrains with high rugosity (e.g., $R_g = 0.32$, Fig.~6.B), local variations in surface roughness are inevitable. As shown in Fig.~6.B, the terrain rendering illustrates regions that are relatively “simple” (exhibiting smaller local height variance) and others that are more “complex” (with larger local height variance). Given that bac generation is sensitive to these local height variations, we propose leveraging CI to adapt the coding scheme dynamically in response to the estimated local rugosity. Notably, instead of applying corrections at the level of individual bacs, we adapt the coding scheme (i.e., the self-propulsion patterns) across the entire robot. In this section, we will evaluate the effectiveness and robustness of this centralized adaptation strategy for locomotion across noisy landscapes.

We first identify how modulation of the coding scheme can serve as an appropriate response to the estimated local rugosity. As discussed in the previous section, the use of a vertical wave introduces a trade-off between advantages (more uniform distribution of thrust generation) and disadvantages (fewer legs making ground contact), which affects locomotion performance over rugose terrains. Thus, the optimal vertical wave amplitude is typically terrain-dependent (see SI). A fixed vertical wave amplitude may perform well in some regions of a complex terrain while under-performing in others. Specifically, at lower values of $A_p$, the robot exhibits higher speed in ``simple” regions but struggles in more complex regions; whereas at higher $A_p$, the robot maintains robust performance in complex regions but moves at lower speed in ``simple" (or intermediate) regions. To illustrate this, we evaluate the open-loop performance of a 12-legged MER on a high-rugosity terrain (Fig.~6.E) and observe significant variance in average speed across different $A_p$ values. Furthermore, the ``optimal" $A_p^*$ empirically measured from higher rugose terrain (Fig.~6.E) also deviates from the one measured from lower rugose terrain (Fig.~5.B). To address this, we propose adapting $A_p$ in response to the estimated local rugosity.

We next identify what sensors can accurately estimate the local rugosity. Inspired by the concept that local rugosity correlates well with bac loss frequency, we use contact sensors (located on the feet) to detect the extent of bac loss. By comparing the desired bac pattern with the measured bac pattern, we can estimate the local terrain rugosity and thereby facilitate the adaptation of the vertical wave. In Fig.~6.D, we illustrate a comparison between the desired bac and the terrain-contaminated bac.

Finally, we construct a simple single-input-single-output linear controller to adaptively choose the vertical wave amplitude (Fig.~6.C). Specifically, we measured the terrain-disturbed duty factor ($\hat{d}$, the fraction of a bac over a period, averaged over all limbs) and choose the vertical wave amplitude, $A_p$, according to the difference between $\hat{d}$ and the desired duty factor ($d$): $A_p = p(d-\hat{d})$. Here, we set $p=3.2$ (with the units of rad). The value of $p$ is chosen from our empirical experiments. To further simply our control and reduce bandwidth, we update $A_p$ once at the end of every period. The real-time recording of $\hat{d}$ and $A_p$ are illustrated in Fig.~6.D.

The use of adaptive vertical wave not only improves the average velocity, but also reduces the variance (Fig.~6.E, red rectangular box). We further illustrate the displacement profile (the average and standard deviation of displacement plotted a function of time) in Fig.~6.F. We compare the  displacement profile for open-loop experiments subjected to different vertical wave amplitude ($A_p=\{0, 2\pi/9\}$), which highlight the importance to use vertical wave to improve average velocity. However, the variance also increase with the vertical wave. The adaptive vertical wave serve to reduce the variance of the locomotion performance. We notice that even after 10 periods, the displacement profile is still tightly bounded, indicating its reliable performance on rugose terrains. 

\subsection*{Approach the upper bound of speed on noisy landscapes}

\subsubsection*{Multi-input-multi-output CI via reinforcement learning}

In this section, we explore more advanced forms of CI to enable effective locomotion of MERs on noisy terrains, with the goal of approaching the theoretical upper bound on speed.
In the previous section, we explored a single-input-single-output (SISO) CI strategy to enhance MER locomotion. Specifically, we modulated only the vertical wave amplitude in response to changes in local rugosity. However, as previously discussed, multiple parameters can influence the trade-off between speed and robustness, such as the lateral body undulation amplitude ($A_b$) and leg amplitude ($A_{\text{leg}}$). In pursuit of improved performance, we posit that a multi-input-multi-output (MIMO) controller could further enhance MER locomotion by leveraging these additional control parameters.

In this section, we explore the implementation of a multi-input-multi-output (MIMO) CI framework. The input variables include the current lateral body undulation amplitude $A_b(t)$, leg amplitude $A_{\text{leg}}(t)$, vertical wave amplitude $A_p(t)$, and the average duty factor $\hat{d}$. The output variables are the updated values $A_b(t+1)$, $A_{\text{leg}}(t+1)$, and $A_p(t+1)$ (Fig. 7.A). 

Unlike the SISO CI, which involves only a single tuning parameter $p$, the MIMO CI include a substantially larger parameter space. To identify an effective MIMO CI model, we employ reinforcement learning techniques to numerically optimize the multilayer perceptron (MLP, Fig.~7.A.1) in control parameters (Fig.7.A). For details on the machine learning model and tuning process, we refer readers to\cite{he2024learning}. The resulting MIMO CI is implemented on a 12-legged MER and evaluated on rugose terrain. As shown in Fig.~7.A.2, the experiments demonstrate a substantial speed improvement when transitioning from SISO CI to MIMO CI. Notably, over noisy landscapes, MIMI CI enables a locomotion speed of $0.40\pm0.03$ BL/cycle, approaching the theoretical upper bound of 0.43 BL/cycle predicted in previous sections.\footnote{The robot used in this section is equipped with contact sensors at the tips of each foot, effectively increasing leg length and thereby raising the upper bound on achievable speed.}

\subsubsection*{Embedding MI in bac through foot morphology design}

In the above sections, we used coding and CI schemes to improve performance on a MER with feet modeled after  biological centipede. These simple limbs, which we  refer to as point legs, make point contact with the substrate and resemble the leg designs of many existing legged robots~\cite{raibert2008bigdog,hutter2016anymal}. Due to the sensitivity of point contacts to local terrain rugosity, robots with point legs often require complex CI algorithms or excessively redundant legs to traverse even moderately uneven terrains effectively.  However, previous research on hexapods~\cite{spagna2007distributed} indicates that in certain rugose terrains, a C-shaped leg can improve performance in open loop  (MI) with limited CI. The key mechanism is that C-legs act as a ``distributed foot" and enable surface contact with the substrate (determined by the arc length and the width of the C-leg). In this way, a bac loss is only considered when the entire C-leg loses contact with the substrate; partial contact still preserves the capability to generate a bac (e.g., Fig.~7.B.1, bac generation from partial surface contact). 
This results in more reliable contact compared to point legs, an evolution also seen in RHex~\cite{altendorfer2001rhex,moore2002leg,jun2014compliant}.

We thus hypothesized that, on certain terrains, C-legs on the MER could allow performance with MI alone comparable to CI plus point leg.  We tested this hypothesis  by equipping MERs with C-shaped legs with arc lengths comparable to the characteristic terrain length scale (e.g., height variation) and evaluating their performance on noisy landscapes (Fig.~7.B.1). We compared locomotion speeds across flat, low-rugosity ($R_g = 0.17$), and high-rugosity ($R_g = 0.32$) terrains and found the performances to be comparable, with no statistically significant differences (Fig.~7.B.3). This is further supported by the displacement profiles over time (Fig.~7.B.2), where the trajectory on the high-rugosity terrain (blue curves) closely resembles that on flat ground (orange curves). Finally, the observed speed approaches the theoretical upper bound (4$\times$ leg length, 0.44 BL/cycle), confirming the effectiveness of the C-shaped leg design.

\subsection*{Demonstration: reliable MER locomotion in laboratory and field complex environment}

We demonstrate the effectiveness of our proposed multi-legged locomotion framework in both laboratory and natural complex environments. Specifically, we tested lab-constructed MERs in real-world terrains (Fig.1.D.1), where the integration of CI with moderate leg redundancy (12 point legs) enabled robust locomotion across diverse and challenging substrates, including grass, pebbles, and inclined surfaces. Additional details on the field tests are provided in\cite{he2024control}.

In addition to CI, we demonstrate that MI in MERs can also aid effective traversal of confined environments in practical scenarios. We present a robot named SCUTTLE (Slithering Centipede-like Undulatory Terradynamically Tactical Legged Explorer), which is currently being commercialized by Ground Control Robotics (GCR). SCUTTLE features moderate leg redundancy (10 C-shaped legs) and operates without any computational intelligence. Instead, it is tele-operated using a set of simple commands: forward, turn left/right, backward, climbing, sidewinding, and self-righting. The forward locomotion mode is described in this paper, and we refer readers to~\cite{flores2024steering,iaschi2024addition,iaschi2024addition} for further details on the other modes. SCUTTLE’s performance was evaluated in the ICRA 2025 Quadruped Robot Challenge~\cite{ICRA25}, where it demonstrated remarkable mobility with minimal CI (see SI video). We also validated its performance in real-world scenarios, such as pipe inspection and traversal over granular substrates like sand and leaf litter, further highlighting its robust, terrain-agnostic mobility with strong potential for field deployment.

\section*{Discussion and conclusion}

In this paper, we establish a connection between theories and algorithms in signal transmission and passive and/or active control strategies for multi-legged locomotion. Specifically, we identify a mechanical intelligence (MI) of MERs as the gravity-induced passive response to terrain uncertainties. We further formulate this MI as analogous to the ``majority vote decoder” in signal transmission, with the corresponding “simple repetition encoder” represented by spatial synchronization control scheme. While this spatial synchronization enhances robustness, it also results in reduced speed in noise-free environments. Using geometric mechanics (GM), we encode for MERs and identify a trade-off spectrum between speed and robustness. Through theoretical, numerical, and experimental analyses, we determine the upper bound of absolute speed achievable in noise-free environments as a function of the number of legs.

To explore beyond the limit of MI-only MERs, we next seek active feedback schemes which we identify as analogous to automatic repeat request (ARQ) in signal transmission. In our context, we leverage computational intelligence (CI) to simultaneously improve both speed and robustness. In particular, we use vertical wave to control the ``majority vote window size" and use a single-input-single-output linear feedback controller to select the appropriate vertical wave amplitude based on simple binary contact sensors. We demonstrate the potential of our proposed framework with a lab-constructed robot in field tests and a robot platform being commercialized by Ground Control Robotics, Inc.

We notice that even with a simple single-input-single-output linear controller, we can significantly improve the performance of a 12-legged MER to a level similar to that of a 16-legged MER without feedback~\cite{chong2023shannon}. This indicates a possible equivalency between computational complexity (e.g., sensors and processors) and design complexity (e.g., more legs). In early work dating back two decades, a two-dimensional control architecture was proposed to prescribe robotic and biological locomotion on complex terrains, comprising two key axes~\cite{koditschek2004mechanical}. The first axis concerns the interplay between feedback and feedforward systems (sensor complexity). The second axis pertains to centralized and decentralized control, indicating how sensor feedback is integrated and processed~\cite{butler2002generic,yasui2019decoding,gilbert1997visual}. Over the past two decades, various centralized and decentralized feedback models have been used to reconstruct animal behaviors~\cite{ijspeert2020amphibious,sponberg2017emergent} (processor complexity). Our framework essentially adds a third axis of design complexity to the classic control architecture. We hypothesize that there should exist an iso-height contour where all points have similar performance, trading off among sensor, processor, and design complexity. Exploration of the 3-dimensional control architecture can benefit soft and micro robots in applications such as search-and-rescue and agriculture, where potential compromised capacity in sensors and processors can be compensated by design complexity.

Finally, beyond advanced CI, we demonstrate that the speed upper bound can also be approached through terrain-specific MI, such as the C-leg design (in the GCR robot SCUTTLE). Just as advanced CI enables robots to adapt their self-deformation patterns in response to terrain conditions, it is promising to explore an additional layer of adaptability by dynamically modulating MI—for example, through real-time shape-morphing of foot geometries~\cite{baines2022multi}. This embodied, hierarchical integration of CI and MI holds the potential to enable terrain-agnostic, agile, and resilient robotic systems capable of robust operation in unstructured and extreme environments.

\section*{Materials and Methods}

\subsection*{Rugose terrains}

To systematically emulate rugose terrains, we used stepfields in accordance with NIST standards for assessing search-and-rescue robot capabilities~\textit{(32)}. Each block is a 10 by 10 cm square with a height between 0 and 12 cm in 1 and 2.5 cm increments for the terrains with rugosity $R_g$ = 0.17 and $R_g$ = 0.32, respectively. The number of blocks associated with each height was determined from a normal distribution generated in MATLAB. We obtained two terrains with a mean and standard deviation of 6.0 and 2.0 cm for the $R_g$ = 0.17 terrain and 6.25 and 4 cm for the $R_g$ = 0.32 terrain. We truncated these distributions between 0 and 12 cm using MATLAB's truncate() command such that we avoided negative heights in our model and extreme heights when physically constructing these terrains. We formed these blocks out of foam (FOAMULAR Insulating Sheathing (IS) XPS Insulation) and laid them spatially across a 2D grid of size $W,H$ where $(W,H)$ = (80, 160) cm for the $R_g$ = 0.17 terrain and (50, 300) cm for the $R_g$ = 0.32 terrain.

\subsection*{Robot experiment protocol and data analysis}

The OptiTrack motion-tracking system was utilized to record the positions and postures of the robot in the workspace. Four motion capture cameras (OptiTrack Prime$^{\text{x}}$ 13) were mounted above the rugose terrain to capture the real-time 3D positions of reflective markers attached to the robot's body. The markers were placed at each joint. The X, Y, and Z position values of each marker were obtained from the Motive software using MATLAB. In addition, a high-resolution camera (Logitech HD Pro Webcam C920) was mounted above the experiment environment to record videos of each experiment. 

Robot experiments consisted of a series of trials running the robot in the rugose terrains. We placed the robot starting at a random initial position for each trial. Depending on the experiments, we stopped the experiments if (1) the robot traverse the entire terrain and is about to step out or (2) the robot have finished the designated task (either over 60~$cm$ in Fig. 2 or finished three cycles of locomotion in Fig. 3-7). 

\section*{Acknowledgment}

We would like to thank the Institute for Robotics and Intelligent Machines for use of the College of Computing basement as a testing space. Additionally, we thank Ground Control Robotics LLC. for use of their robotic platform and support. 
The authors received funding from NSF-Simons Southeast Center for Mathematics and Biology (Simons Foundation SFARI 594594), Army Research Office grant W911NF-11-1-0514, Georgia Research Alliance (GRA) AWD-005494, GA AIM AWD-004173, a Dunn Family Professorship, Ground Control Robotics, and a STTR Phase I (2335553) NSF grant.

\clearpage
\begin{figure*}[ht]
\centering
    \includegraphics[width=0.70\linewidth]{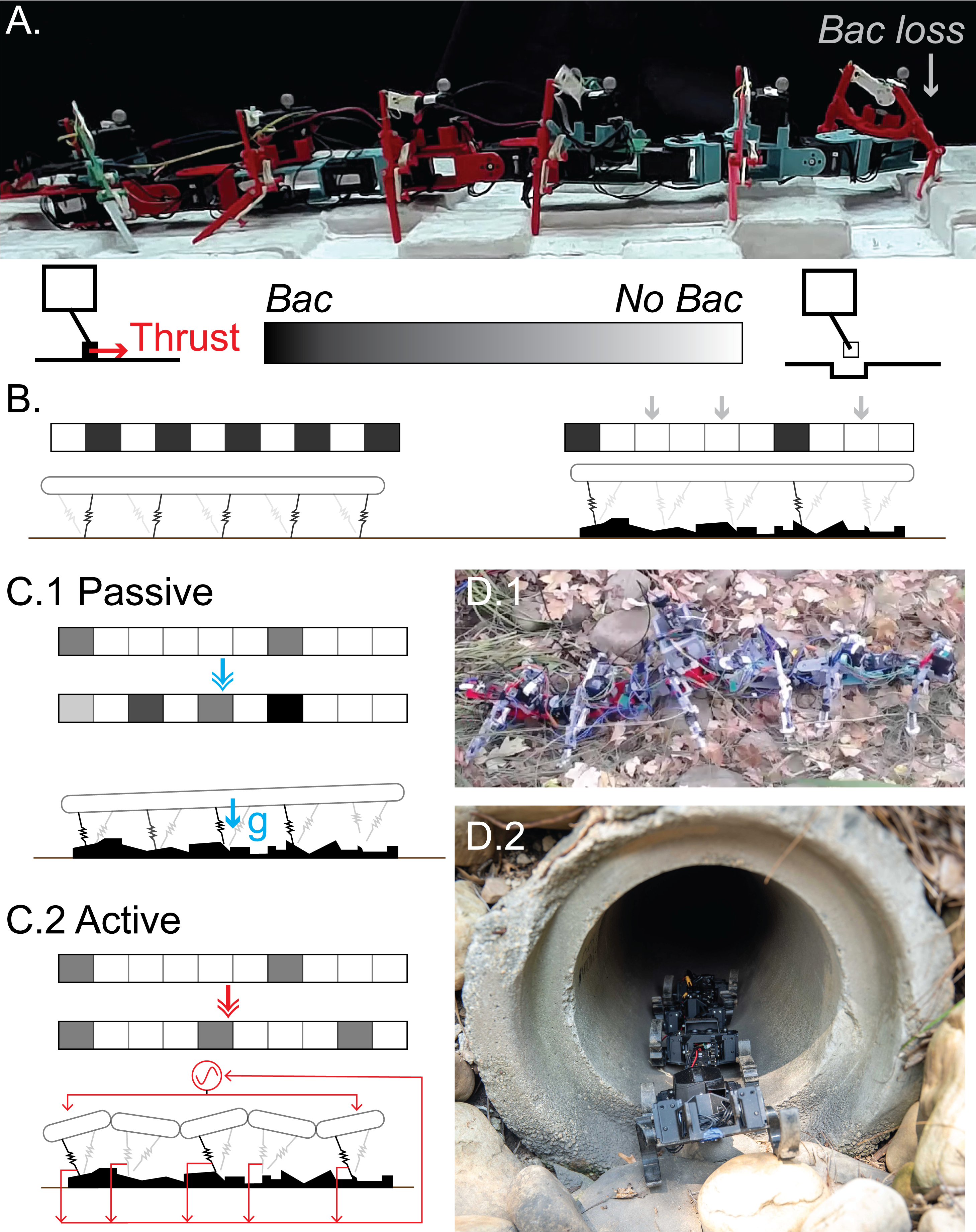}
    \caption{\textbf{Active and passive coding for MERs} (A) (\textit{top}) A side-view picture of a 10-legged MER locomoting on a rugose terrain. White arrow points to the leg with a bac (basic active contact) loss. (\textit{bottom}) An illustration of (\textit{left}) a bac and (right) a terrain-induced bac loss. (B)  An illustration of the 10-legged MER moving on (\textit{left}) flat and (\textit{right}) rugose terrains. Uneven terrains will introduce noise to the bac pattern. The instantaneous bac pattern is illustrated on the top panel. (C.1) Gravity will re-distribute the ground reaction forces among all the bac, serving as an effective ``majority vote" decoder. (C.2) Active feedback control to counter the terrestrial perturbation via vertical wave adaptation. (D.1)  A 12-legged lab-constructed MER moving on natural complex environment. (D.2) Demonstration of MER agility using a commercial robot platform (Ground Control Robotics, Inc).}

\end{figure*}

\begin{figure*}[ht]
\centering
    \includegraphics[width=0.70\linewidth]{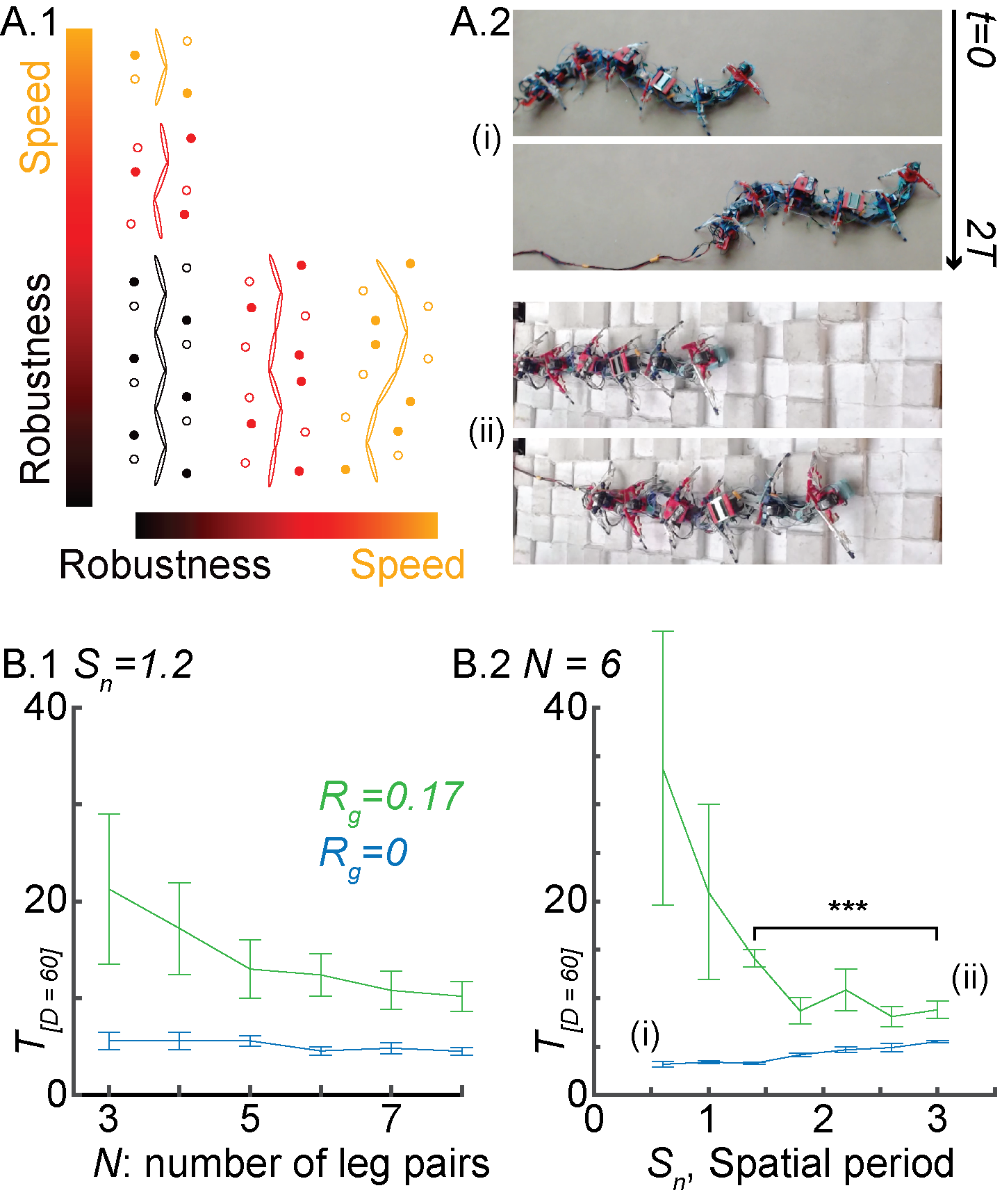}
    \caption{\textbf{Trade-off between robustness and speed} (A.1) Column shows the trade-off between speed and robustness as a function of design redundancy for a fixed control. Row shows the trade-off between speed and robustness  as a function of control redundancy for a given robot. (A.2) Snapshots of a 12-legged MER with (i) 0.6 and (ii) 3 spatial periods moving flat and rugose terrains respectively. (B.1) $T_{[D=60]}$, time required to travel 60$cm$ ($\pm$ standard deviation over 10 trials) as a function of design redundancy ($N$, the number of leg pairs) on flat (blue curve) and rugose (green curve) terrains (adapted from~\cite{chong2023multilegged}). (B.2) $T_{[D=60]}$ as a function of control redundancy ($S_n$, spatial period) on flat and rugose terrains. }
\end{figure*}

\begin{figure*}[ht]
\centering
    \includegraphics[width=0.40\linewidth]{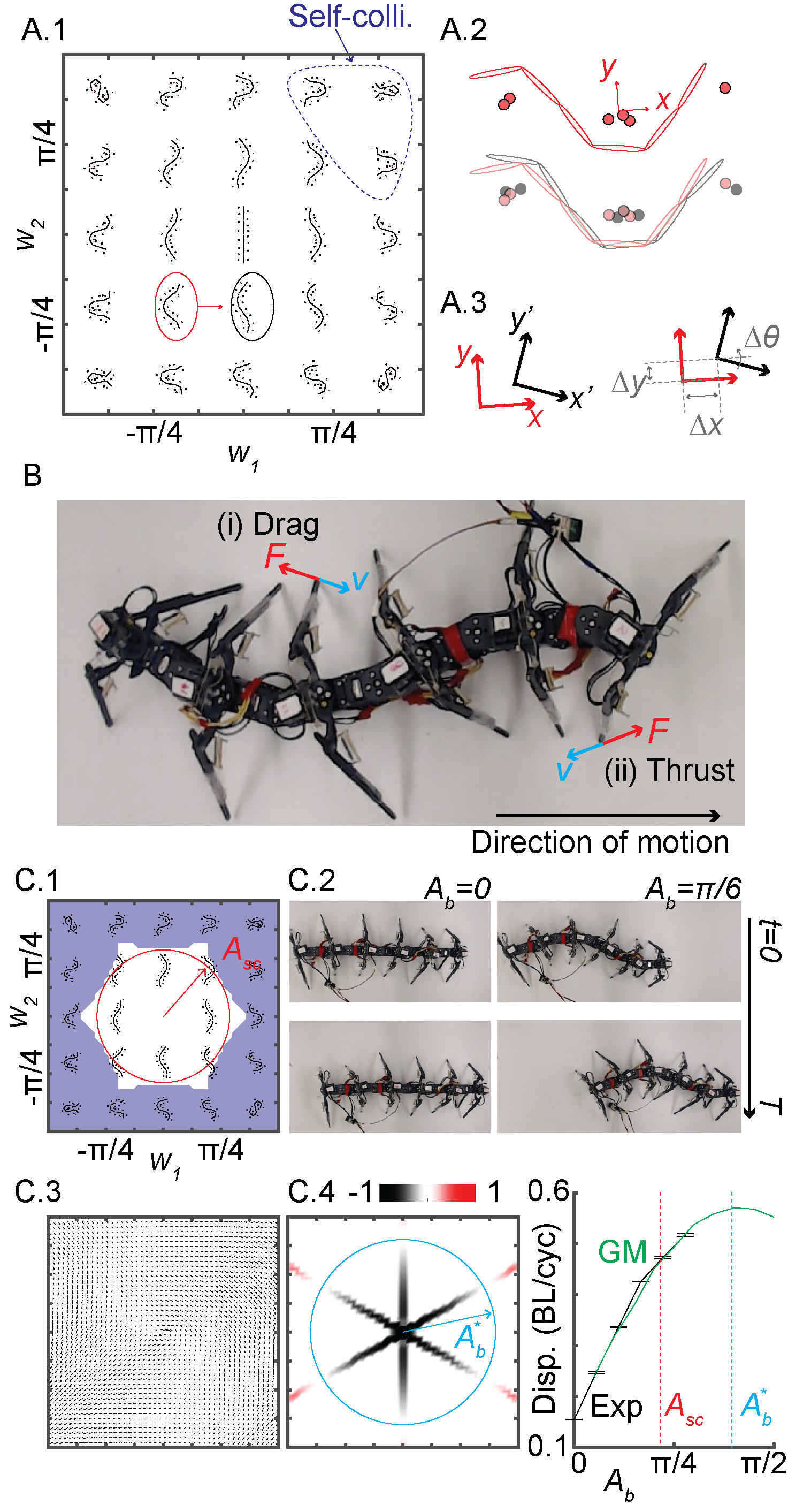}
    \caption{\textbf{Geometry of MER locomotion} (A.1) The shape space for body undulation of a 12-legged MER. Each point in the shape space prescribes a specific body posture (including body/leg joint angles and contact distribution). (A.2) (\textit{top}) One body posture in the shape space, corresponding to the red circle in A.1. (\textit{bottom}) Shape changes (from the body posture circled in red to the posture circled in black) can lead to net translation and rotation. (A.3) The illustration of net translation ($\Delta x$ and $\Delta y$) and rotation ($\Delta \theta$) as a function of shape changes. (B) An illustration of isotropic Coulomb friction where the ground reaction force ($F$) aligns with the opposite direction of foot slipping ($v$). (C.1) The shape space of a 12-legged MER with self-collision shapes marked in shaded blue color. Red path denotes the largest circular path (radius: $A_{SC}$) in the shape space without self-collision. (C.2) Snapshots of a 12-legged MER with different amplitudes of lateral body undulation moving on flat ground over one cycle. (C.3) The vector field, $\bA^x(\dw)$, of a 12-legged MER.  (C.4) (\textit{left}) The forward height function of a 12-legged MER. The blue path denotes the circular gait path (radius: $A_{b}^*$) enclosing the most surface integral over the forward height function. (\textit{right}) Comparison of GM predicted (green) and experimentally measured (black) velocity (units: body length per cycle) as a function of $A_{b}$. $A_{SC}$ and $A_{b}^*$ are labeled as dashed lines.}
\end{figure*}

\begin{figure*}[ht]
\centering
    \includegraphics[width=0.50\linewidth]{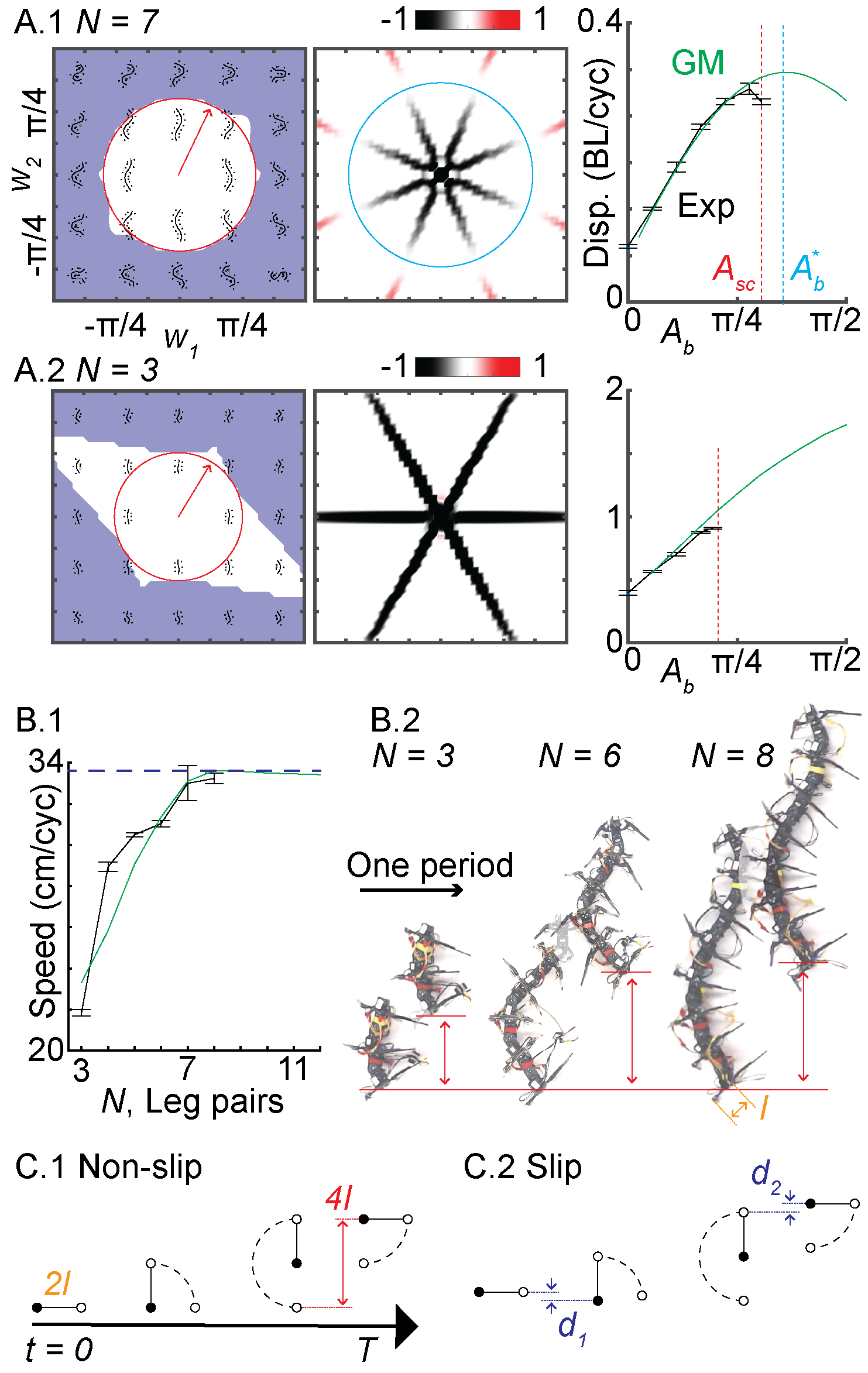}
    \caption{\textbf{Upper bound on absolute velocity on flat ground} (A) Body undulation analysis for (A.1) a 14-legged MER and (A.2) a 6-legged MER respectively. (\textit{left}) Shape space with numerically computed self-collision region labeled in blue shade. (\textit{mid}) The forward height function. (\textit{right}) Comparison between geometric mechanics and robot experiments. For a 6-legged MER in (A.2), $A_{b}^*>\pi/2$ and thus not illustrated in the figure. Axes are identical for all shape spaces. (B.1) the absolute velocity (unit: cm per cycle) as a function of $N$. Geometric mechanics prediction and robot experiments are compared. The blue dashed line represents the theory-predicted upper bound. (B.2) Snapshots of (\textit{left}) a 6-legged, (\textit{mid}) a 12-legged, and (\textit{right}) a 14-legged MER locomotion on flat ground with the optimal feasible body amplitude. The leg length, $l$ is labeled. (C) Characterizatio of the upper bound on the absolute velocity from (C.1) non-slip and (C.2) slipping models.}
\end{figure*}

\begin{figure*}[ht]
\centering
    \includegraphics[width=0.60\linewidth]{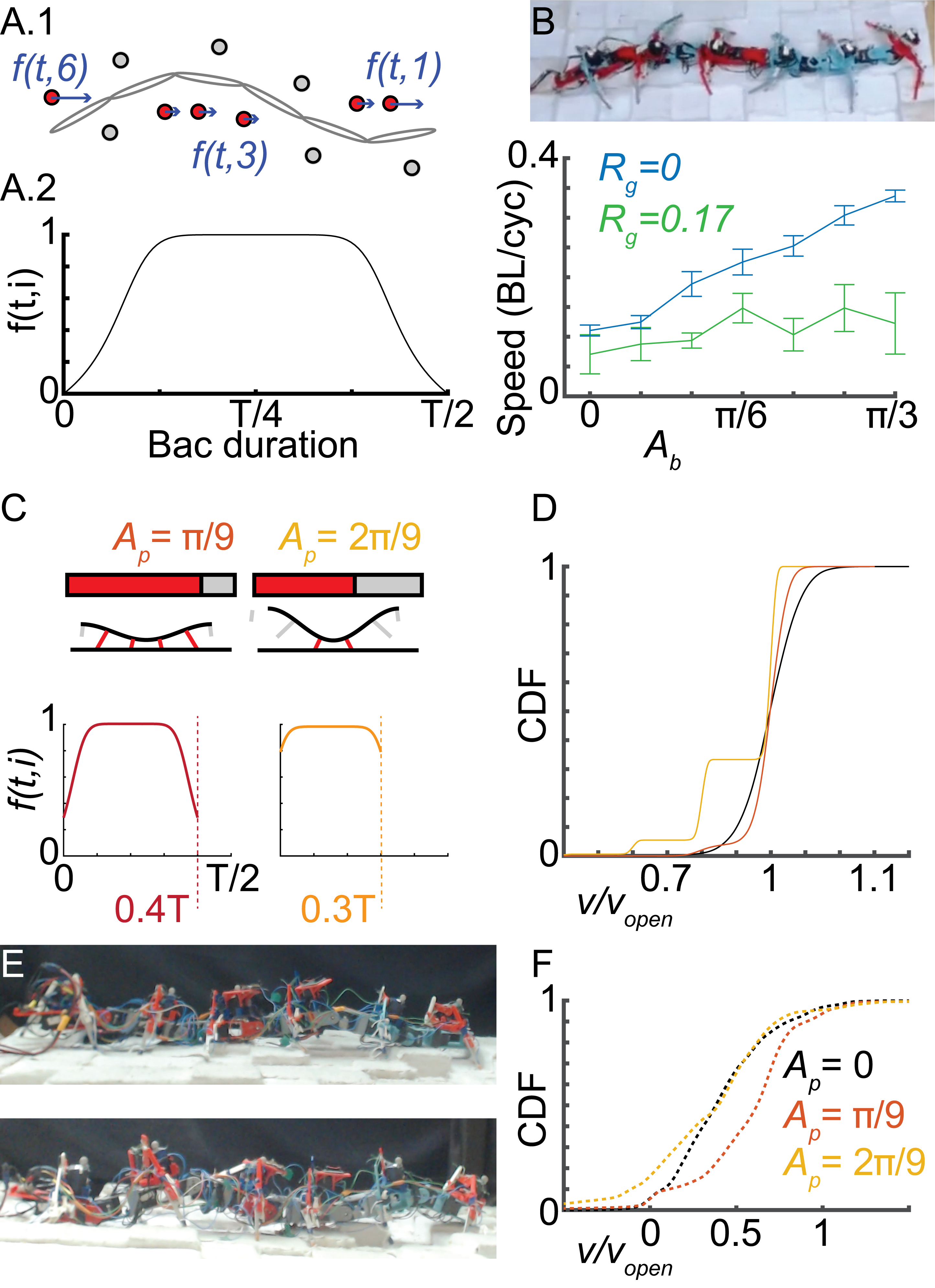}    
    \caption{\textbf{Contact modulation to counter terrain rugosity} (A.1) An illustrations of thrust generation from bacs (basic active contact). (A.2)  The instantaneous thrust (unit: $\mu N$, $\mu$ is friction coefficient and $N$ is the normal supporting force) as a function of time, derived from~\cite{chong2023self}. 
    (B) (\textit{top}) Top-view of a 12-legged MER on rugose terrain.   (\textit{bottom}) Experimentally measured average velocity on a flat terrain (blue curve) and a rugose terrain ($R_g=0.17$, green curve) as a function of lateral body undulation amplitude $A_b$. 
    (C) (\textit{Top}) An illustration of contact modulation via vertical body undulation. (\textit{Bottom}) Vertical wave modulations lead to a more uniform distribution of instantaneous thrust function but with fewer bac duration. (D) The numerically calculated cumulative distribution function (CDF) of normalized terrain-disturbed average velocity ($v/v_{open}$) subject to different contact modulations. (E) Snapshots of a 12-legged MER locomote on a rugose terrain with (\textit{top}) no vertical body undulation and (bottom) $A_p = 2\pi/9$. (F) CDF of experimentally measured cycle-average velocity $v/v_{open}$ on the rugose terrain subject to different $A_p$ are compared.
    }
\end{figure*}

\begin{figure*}[ht]
\centering
    \includegraphics[width=0.60\linewidth]{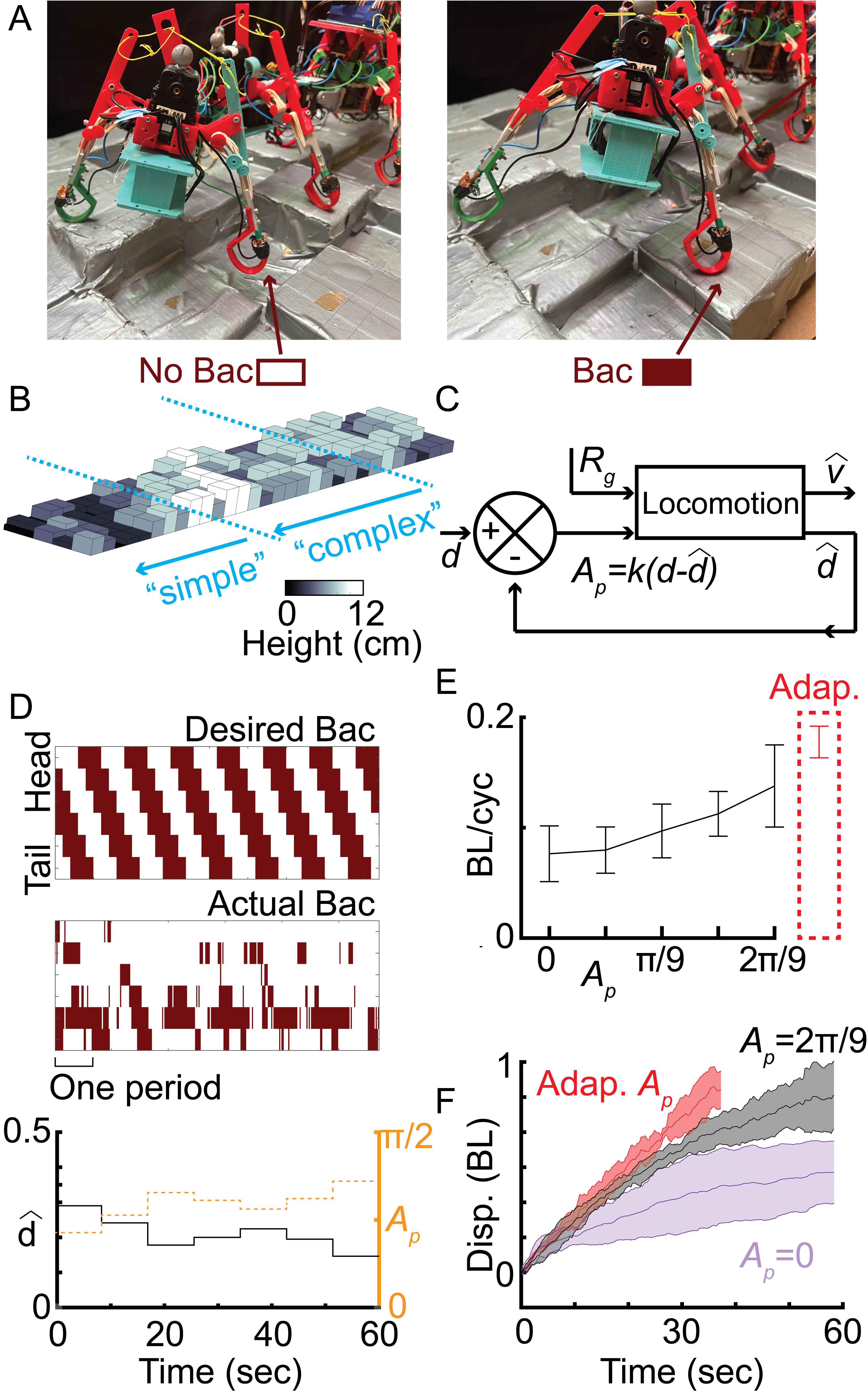}    \caption{\textbf{Active coding to improve locomotion performance on terrains with varying rugosity.}   (A) The use of a contact sensor to detect bac loss. (B) Rendering of a rugose terrain with ``simple" and ``complex" parts of the terrain labeled.   (C) The simple linear single-input-single-output controller. (D) (\textit{Top}) The comparison of the desired bac sequence and the actual detected bac sequence over 7 periods. (\textit{Bottom}) Black curve denotes the recorded average bac duration in each period. The orange dashed curve denotes the adaptive vertical wave amplitudes. (E) (\textit{Top}) Experimental measured average velocity (unit: BL/cyc) of the robot on higher rugose terrains subject to different vertical wave amplitudes ($A_p$). The adaptive vertical wave performance is compared on red rectangle box.    (F) The comparison of displacement profiles (units: BL) for (purple) $A_p=0$, (black) $A_p=2\pi/9$, and (red) adaptive $A_p$.
    }
\end{figure*}

\begin{figure*}[ht]
\centering
    \includegraphics[width=0.60\linewidth]{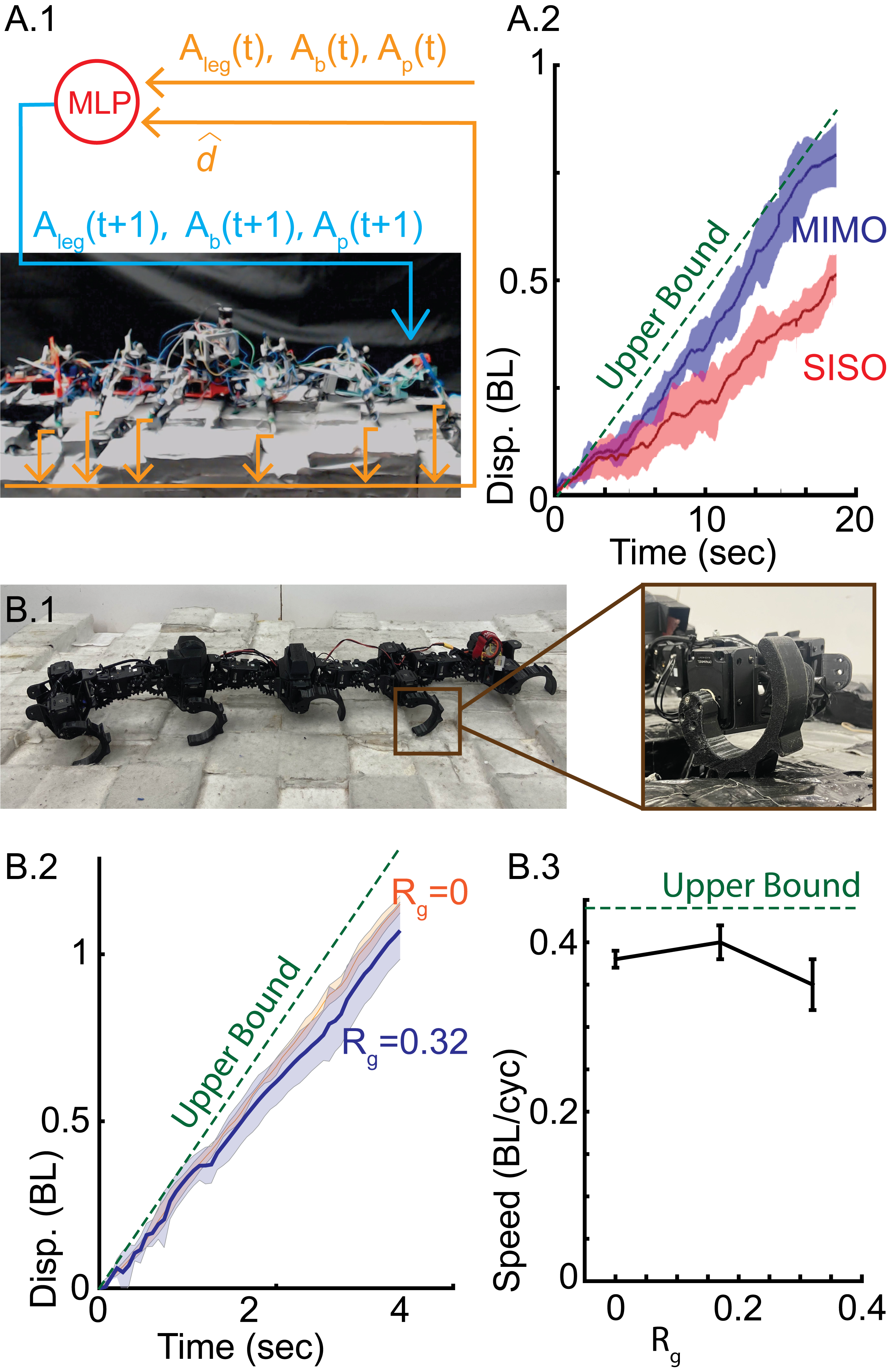}    \caption{\textbf{Leaning multi-input-multi-output (MIMO) CI for MER locomotion.} (A.1) The MIMO CI model. The input variables are current body lateral and vertical amplitude ($A_b(t)$ and $A_p(t)$), current leg amplitude ($A_{leg}(t)$), and the measured average duty factor ($\hat{d}$). The output variables are updated $A_b(t+1)$, $A_p(t+1)$, and $A_{leg}(t+1)$. We use reinforcement learning to fine tune the multilayer perceptron (MLP) for MIMO CI. (A.2) The comparison of displacement profiles (units: BL) for speed upper bound (dash green), (blue) MIMO, and (red) SIMO CI. MIMO CI is approach the theoretical upper bound of speed. (B.1) (\textit{left}) C-leg MER on rough terrain. (\textit{right}) A snapshot of C-legged MER interaction with substrate. B.2. The comparison of displacement profiles (units: BL) for speed upper bound (dash green), (blue) on higher rugose terrain ($R_g = 0.32$), and (orange) flat terrain ($R_g = 0$). (B.3) Experimentally measured average speed of C-leg MER plotted as a function of terrain rugosity. Error bar represents the standard deviation.}
\end{figure*}

\clearpage
\bibliography{mainbib.bib}
\bibliographystyle{Science}
\end{document}